\documentclass[11pt,fa4paper]{article}
\usepackage[nohyperref]{acl2020}

\usepackage{times}
\usepackage{latexsym}
\usepackage{microtype}
\usepackage{amsfonts, amsmath, amsthm}
\usepackage{breqn}
\usepackage{makecell}
\usepackage{scalerel}
\usepackage{subfig}
\usepackage{xspace}
\usepackage{url}
\usepackage{xfrac}
\usepackage{xspace}
\usepackage{booktabs}
\usepackage{tabstackengine}
\usepackage{adjustbox}
\usepackage{longtable}

\usepackage{tikz}
\newsavebox{\tempbox}
\usepackage{arydshln}
\usepackage{enumerate}
\usepackage[inline]{enumitem}
\usepackage{floatrow}
\usepackage{cleveref}
\crefname{section}{\S}{\S\S}
\crefname{table}{Table}{}
\crefname{figure}{Figure}{Figures}
\crefname{algorithm}{Algorithm}{}
\crefname{equation}{eq.}{}
\crefname{appendix}{App.}{}
\crefname{prop}{Proposition}{}
\crefname{thm}{Theorem}{}
\crefformat{section}{\S#2#1#3}  %

\newcommand{\vtheta}{{\boldsymbol \theta}}
\newcommand{\KL}{\mathrm{KL}}
\newcommand{\uu}{u}
\newcommand{\q}{(1-\alpha)u + \alpha p}

\newcommand{\Jalphaf}{\mathrm{J}_{\alpha,G}}
\newcommand{\Jalphah}{\mathrm{J}_{\alpha,-\ent}}
\newcommand{\Jalpha}{\mathrm{J}_{\alpha}}

\newcommand{\Jone}{\mathrm{J}_{1}}
\newcommand{\Jzero}{\mathrm{J}_{0}}
\newcommand{\Jhalf}{\mathrm{J}_{1/2}}
\newcommand{\JS}{\mathrm{JS}}
\newcommand{\nabtheta}{\nabla_{\scaleto{\vtheta}{5pt}}}
\newcommand{\ptheta}{p_{\scaleto{\vtheta}{4pt}}}
\newcommand{\calX}{{\cal X}}
\newcommand{\calC}{{\cal C}}

\newcommand{\calYs}{{\cal Y}}
\newcommand{\spq}{{(1-\alpha)p + \alpha q}}
\newcommand{\xx}{\mathbf{x}}
\newcommand{\yy}{\mathbf{y}}
\newcommand{\ptilde}{\tilde{p}}

\newcommand{\defeq}{\mathrel{:\mkern-0.25mu=}}
\newcommand{\ent}{\mathrm{H}}
\newcommand{\ra}[1]{\renewcommand{\arraystretch}{#1}}
\newcommand{\calL}{\mathcal{L}}
\newcommand{\bleu}{\textsc{bleu}\xspace}
\newcommand{\reg}{\mathcal{R}}
\newcommand{\eos}{\textsc{eos}\xspace}
\newcommand{\bos}{\textsc{bos}\xspace}
\newcommand{\defn}[1]{\textbf{#1}}

\newtheorem{prop}{Proposition}

\definecolor{darkgreen}{RGB}{0,160,0}
\newcommand{\dd}[1]{\textcolor{darkgreen}{\bf\footnotesize +#1}}

\usepackage[section]{placeins}

\usepackage{tipa}

\newcommand{\ucambridge}{ \textrm{\normalfont \textipa{D}}}
\newcommand{\ethz}{\textrm{\normalfont \textipa{Q}}}
\newcommand{\jhu}{\textrm{\normalfont \textipa{Z}}}
\aclfinalcopy 
\def\aclpaperid{1001} 

\everypar{\looseness=-1}

\title{Generalized Entropy Regularization or: \\ There's Nothing Special about Label Smoothing}

\author{Clara Meister$^\ethz$~\;~Elizabeth Salesky$^\jhu$~\;~Ryan Cotterell$^{\ucambridge,\ethz}$ \\
  $^\ethz$ETH Z\"{u}rich~\;~$^\jhu$Johns Hopkins University~\;~$^\ucambridge$University of Cambridge \\
  \texttt{clara.meister@inf.ethz.ch}~\;~\texttt{esalesky@jhu.edu} \\ \texttt{ryan.cotterell@inf.ethz.ch}
}
\begin{document}
\setlength{\belowdisplayskip}{8pt}\setlength{\belowdisplayshortskip}{8pt}
\setlength{\abovedisplayskip}{8pt} \setlength{\abovedisplayshortskip}{8pt} 

\maketitle
\begin{abstract}
    Prior work has explored directly regularizing the output distributions of probabilistic models to alleviate peaky (i.e. over-confident) predictions, a common sign of overfitting. This class of techniques, of which label smoothing is one, has a connection to entropy regularization. Despite the consistent success of label smoothing across architectures and datasets in language generation tasks, two problems remain open: (1) there is little understanding of the underlying effects entropy regularizers have on models, and (2) the full space of entropy regularization techniques is largely unexplored. We introduce a parametric family of entropy regularizers, which includes label smoothing as a special case, and use it to gain a better understanding of the relationship between the entropy of a trained model and its performance on language generation tasks. We also find that variance in model performance can be explained largely by the resulting entropy of the model. Lastly, we find that label smoothing provably does not allow for sparse distributions, an undesirable property for language generation models, and therefore advise the use of other entropy regularization methods in its place.
    Our code is available online at \url{https://github.com/rycolab/entropyRegularization}.
\end{abstract} 

\section{Introduction}\label{sec:intro}

When training large neural networks with millions of parameters, regularization of some form is needed to prevent overfitting, even when large amounts of data are used;  models for language generation are no exception. In probabilistic modeling, e.g. when the final layer of the neural network is a softmax, overfitting often manifests itself in overconfident placement of most of the probability mass on a few candidates, resulting in peaky (low-entropy) probability distributions over the vocabulary.
Specifically for language generation tasks, this behavior leads to the output of repetitive or frequently occurring but unrelated text, which is detrimental to the generalization abilities of the model \cite{decoding, holtzman2019curious}.
A natural regularizer to consider is, therefore, one that penalizes overconfidence, encouraging higher entropy in the learned distribution. 
Indeed, the literature has ascribed gains of $\approx 1$ \bleu point in machine translation to label smoothing, one such technique \cite{rec_adv_ml}.

Despite the clear relationship between low entropy and overfitting, only a handful of distinct entropy regularizers have been explored. To fill this gap, we introduce \textbf{generalized entropy regularization} (GER), a unified framework for understanding and exploring a broad range of entropy-inducing regularizers. GER is based on the skew-Jensen family of divergences  $\Jalphaf$ \cite{Nielsen_2011} and thus may be generalized to any Bregman divergence through the choice of generator function $G$. 
For the negative entropy generator function, GER recovers label smoothing \cite{label_smoothing} as $\alpha \rightarrow 1$, and the confidence penalty \cite{confidence_penalty} as $\alpha \rightarrow 0$. 
We provide formal properties of GER in \cref{sec:GER}, proving these special-case equivalences among other characteristics of GER. We then use GER to examine the relationship between entropy and the evaluation metrics in two language generation tasks: neural machine translation (NMT) and abstractive summarization.

GER encompasses a large family of regularizers, which allows us to directly compare label smoothing to other forms of entropy regularization. By studying the relationship between different regularizers on the performance of natural language generation (NLG) systems, we can better understand not just \emph{when} but also \emph{why} label smoothing aids language generation tasks. Through our analysis, we gain the following insights:
\begin{enumerate}[label=(\roman*)]
    \item With tuning of the regularizer's coefficient, \emph{any} choice of $\alpha$ can yield similar performance, i.e. there is nothing special about label smoothing. In fact, our results suggest that label smoothing ($\alpha \rightarrow 1$) makes it more difficult to tune the regularizer's coefficient. 
    \item Label smoothing assigns infinite cost to sparse output distributions, which may be an undesirable behavior for language generation tasks.
    \item  There is a strong (quadratic) relationship between a model's performance on the evaluation metric and its (average) entropy, offering a hint as to why these regularizers are so effective for NLG.
\end{enumerate}
In summary, entropy-inducing regularizers are a boon to probabilistic NLG systems, which benefit from higher entropy output distributions.
Label smoothing works \emph{because} it forces the model towards a higher-entropy solution, but we recommend 
the confidence penalty and other entropy regularizers ($\alpha < 1$) for reasons (i) and (ii) above.

\section{Preliminaries}\label{sec:KL}

In this work, we consider conditional probability models $\ptheta(\yy \mid \xx)$ for natural language generation; such models assign probability to a target sequence $\yy \in \calYs$ given a source sequence $\xx$. Specifically, our target sequence $\yy = \langle y_1, \dots, y_n \rangle$ of arbitrary length $n$ is a sequence of target words\footnote{Targets $y_i$ may also be characters or subwords; our experiments use byte-pair encoding \cite{sennrich2016bpeacl}} $y_i$ from our vocabulary $Y$. The set of all complete target sequences, which are padded with distinguished beginning- and end-of-sentence symbols, \bos and \eos, is then defined as $\calYs \defeq \{ \bos \circ \yy \circ \eos \mid \yy \in Y^* \}$. For language generation tasks, $\ptheta(\yy \mid \xx)$ is typically a neural network with parameters $\vtheta$; this network is often trained to approximate $\ptilde(\yy \mid\xx)$, the empirical distribution (i.e. the distribution of the data). Here, we focus on locally normalized models; in such models $\ptheta(\yy \mid \xx)$ is factored as:
 \begin{align}
    \ptheta(\yy\mid \xx) &= \ptheta(y_1 \mid\xx)\cdots \ptheta(y_n \mid\xx, \yy_{<n})
\end{align}
\noindent where $\ptheta(y_i \mid \xx, \yy_{<i})$ is defined by a softmax over the output of the final fully connected layer of the network. 
Generation is performed using greedy search, beam search or a sampling scheme. Of the candidate sequences generated, the one with the highest probability under the model $\ptheta$ is returned as the model's prediction.

One way of selecting the parameters $\vtheta$ is to minimize the KL-divergence between
the empirical distribution and the model. This yields the cross-entropy loss (plus an additive constant):\footnote{$\ent(p, q) \defeq -\sum_{z \in\mathcal Z}p(z)\log q(z)$ is cross-entropy and $\ent(p) \defeq \ent(p, p) = -\sum_{z \in \mathcal Z}p(z)\log p(z)$ is the Shannon entropy, for which $\log = \log_2$ and $\mathcal Z = \mathrm{supp}(p)$.}
\begin{align}
    \mathcal L(\vtheta) &= \KL(\ptilde \mid\mid \ptheta)\\
    &= \underbrace{\ent(\ptilde, \ptheta)}_{\textrm{\tiny cross-entropy loss}}\, -  \underbrace{\ent(\ptilde)}_{\textrm{\tiny constant w.r.t. $\vtheta$}}
    \label{eq:loss}
\end{align}
However, fitting a model that perfectly approximates the empirical distribution is, in general, fraught with problems \cite{hastie01statisticallearning}. The goal of learning is to generalize beyond the observed data. Exactly fitting the empirical distribution, often termed overfitting, is therefore not an ideal goal and for language generation models specifically, does not go hand-in-hand with the ability of a model to generate desirable text \cite{bengio2015scheduled}. Consequently, it is advisable to minimize a regularized
objective to prevent overfitting:
\begin{equation}\label{eq:standard_reg}
     \mathcal{L}(\vtheta) + \beta\, \reg(\vtheta)
\end{equation}
where $\reg(\vtheta)$ is a regularizer defined over the model with ``strength'' coefficient $\beta > 0$.

\begin{table*}[t!]
\setlength\arrayrulewidth{0.3pt}
\ra{1.2}
  \centering
  \setlength\tabcolsep{15pt}
  \adjustbox{width=\textwidth}{
  \begin{tabular}{ @{}lcl@{} }
   \hline
   \textbf{Training Method} & \multicolumn{1}{l}{\qquad\textbf{Loss Function}} & \textbf{ Alternate Formulation} \\
  \toprule
    Cross Entropy &  \multicolumn{1}{l}{~~~~~~$\mathcal L(\vtheta) = \ent(\ptilde , \ptheta) $} &$=\KL(\ptilde  \mid\mid \ptheta) + \ent(\ptilde)$  \\
    \hdashline
    Confidence Penalty, $D_\Jzero$ & ~~$\mathcal L_{\textit{CP}}(\vtheta) =\mathcal L(\vtheta) + \beta\, D_\KL(\ptheta\mid\mid u)$ & $ =\mathcal L(\vtheta) - \beta\, D_\ent(\ptheta) + C$\\
    Label Smoothing, $D_\Jone$ &  ~~$\mathcal L_{\textit{LS}}(\vtheta) =\mathcal L(\vtheta) + \beta\,D_\KL(u \mid\mid \ptheta)$  &  $=\mathcal L(\vtheta) + \beta\,D_\ent(u,\ptheta) + C$ \\
    Generalized Entropy Regularization, $D_\Jalpha$ & $\mathcal L_{\textit{GER}}(\vtheta) =\mathcal L(\vtheta) + \beta\, D_\Jalpha (u\mid\mid \ptheta) $  & \qquad \qquad ~~~~ --- \\
     \bottomrule
  \end{tabular}}
  \caption{Loss functions and their alternate formulations for different training methods; the latter three are entropy regularization techniques that augment the standard loss function in row 1. $C$ denotes a constant with respect to $\vtheta$.
  }
  \label{tab:objectives}
\end{table*}
\subsection{Entropy Regularization}
Overfitting can manifest itself as peakiness in $\ptheta$ \cite{Williams1991FunctionOU, pmlr-v48-mniha16, confidence_penalty}.
In other words, $\ptheta$ overconfidently places
most of the probability mass on very few candidates.
While this overconfidence improves training loss, it hurts generalization. Entropy regularization
is one technique that directly combats such overconfidence by encouraging more entropic (less peaky) distributions. 

The entropy of the model $\ptheta$ is defined as
\begin{equation}
   \ent\left(\ptheta\right)  
    \defeq -\sum_{\yy \in \calYs} \ptheta(\yy) \log \ptheta(\yy)
    \label{eq:entropy}
\end{equation}
where we remove dependence on $\xx$ for notational simplicity.
However, the sum in \cref{eq:entropy} over $\calYs$ generally
renders its computation intractable.\footnote{The notation used by \citet{confidence_penalty} is imprecise.} Instead, regularization is performed on the conditional distribution over $Y \cup \{\eos\}$ at each time step, which can be interpreted as an approximation of the true model entropy. For ease of notation, we define a higher-order function $D_f$ over our training corpus $\calC$ consisting of $\langle\xx,\yy\rangle$ pairs that maps a function $f$ over distributions $p,q$ as follows below:
\begin{align}
    D_f(p \mid\mid q) &=\\ \sum\limits_{\scriptscriptstyle{\langle\xx,\yy\rangle \in \calC}} &\sum_{t=1}^{|\yy|} f(p(\cdot \mid \xx,\yy_{<t})\mid\mid q(\cdot \mid     \xx,\yy_{<t}))\nonumber
\end{align}
The function $D_f$ allows us to describe in notation
how entropy regularization is typically employed in 
the training of language generation systems.\footnote{Note that the standard loss function in \cref{eq:loss} can be written in this form when computed over $\calC$, i.e. $\KL(\ptilde \mid\mid \ptheta) = D_\KL(\ptilde \mid\mid \ptheta)$, since the reference $\yy$ is the only value in $\mathrm{supp}(\ptilde)$.}

\paragraph{Label Smoothing.}
Label smoothing, first introduced as a regularizer for neural networks by \newcite{label_smoothing}, is so named because the technique smooths hard target distributions. One such distribution, the empirical distribution, is encoded as a set of one-hot vectors (hard targets) where for each data point, the correct label (e.g., vocabulary index of a word) has value $1$ and all other labels have value $0$.
Label smoothing with strength coefficient $\gamma$ is an add-$\gamma$ smoothing scheme on the distribution over labels at every time step.
Interestingly, minimizing the cross entropy between this modified distribution and the model $\ptheta$ is equivalent to adding the weighted KL divergence between the uniform distribution and the model $\ptheta$ in our original objective function with the same strength coefficient:\looseness=-1
\begin{equation} \label{eq:LS}
   \calL(\vtheta)^{\textit{LS}}_\gamma \defeq (1 - \gamma)\calL(\vtheta) + \gamma \,D_\KL(u\mid\mid \ptheta)
\end{equation} 

\noindent While the loss function is often scaled as above, it is nonetheless equivalent to $\calL(\vtheta)^{\textit{LS}}_{\beta} = \calL(\vtheta) + \beta \,D_\KL(u\mid\mid\ptheta)$;\footnote{up to multiplicative factor $(1 - \gamma)$ when $\beta  = \gamma/(1 - \gamma)$} we use this form for consistency.

\paragraph{Confidence Penalty.}
The confidence penalty, empirically explored in the supervised learning setting by \citet{confidence_penalty}, aims to penalize a low-entropy model. This is done by subtracting a weighted term for the entropy of the model's prediction $\ptheta(\cdot)$ from the loss function, thereby encouraging a more entropic model. This is equivalent to adding the KL divergence between the model $\ptheta$ and the uniform distribution: 
\begin{equation}\label{eq:CP}
     \calL(\vtheta)^{CP}_\beta \defeq  \calL(\vtheta) + \beta\,D_\KL(\ptheta \mid\mid u)
\end{equation}
While \citet{confidence_penalty} found that label smoothing performed better than the confidence penalty for NMT, they only searched coarsely over a small range of $\beta$'s for both regularizers. Our findings in \cref{sec:exp} suggest an alternate conclusion.

\section{Generalized Entropy Regularization}\label{sec:GER}

The positive effect of both label smoothing and the confidence penalty on model performance in language generation tasks motivates further exploration of entropy-promoting regularizers. To this end, we construct a parameterized family of regularizers with label smoothing and the confidence penalty as special cases. We discuss the formal properties of a subset of this family, providing upper and lower bounds for it. We show divergence only occurs in one case for this subset ($\alpha \rightarrow 1$), which directly implies that no sparse solution exists when label smoothing is used as a regularizer.

\subsection{A Family of Entropy Regularizers}\label{sec:family}

We derive a family of regularizers from the skew-Jensen divergence $\Jalphaf$ \cite{Nielsen_2011}, which is defined below as: 
\begin{align} \label{eq:def}
    \Jalphaf(q \mid\mid \ptheta) \defeq &\frac{1}{\alpha (1-\alpha) }\Big( (1-\alpha)G(q) + \alpha G(\ptheta) \nonumber \\ 
    & -  G((1-\alpha)q + \alpha \ptheta) \Big) 
\end{align}
for a strictly convex generator function $G: \Omega \xrightarrow{} \mathbb{R}$ and $\alpha \in (0,1)$ where $\Omega$ is a closed convex set. In this paper, we restrict $\Omega$ to be the $(|Y|+1)$-simplex.
Note that $\Jalphaf(q \mid\mid \ptheta) \neq \Jalphaf(\ptheta \mid\mid q)$ in general, although this is true for some choices of $G$ and $\alpha$.

We define the generalized entropy regularizer as $\reg(\vtheta) = D_\Jalphaf(u \mid\mid\ptheta)$
where $u$ is the uniform distribution.\footnote{Distributions other than $u$ may also be used. See \cref{sec:dist}.} These regularizers promote entropy because they push the model $\ptheta$ towards $u$, which is the maximum-entropy distribution with an entropy of $\log(|Y|+1)$.
Throughout the rest of this paper, we primarily use the generator function\footnote{We also experiment with $G(z) = ||z||_2^2$.} $G(p) = -\ent(p)$. We use $\Jalpha$ as shorthand for $\Jalphah$.\looseness=-1

\begin{figure}
\centering
    \includegraphics[width=0.9\textwidth]{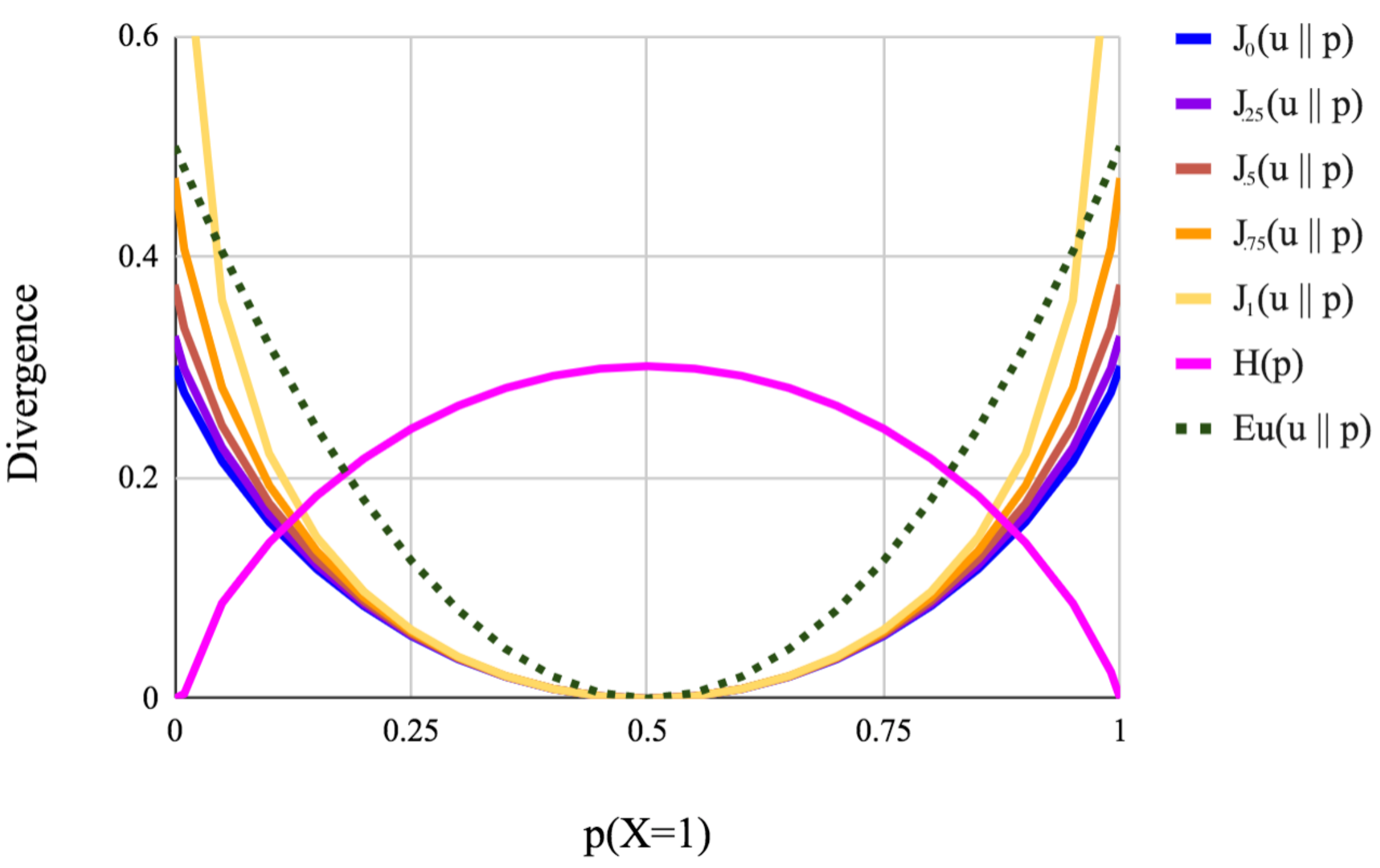}
  \caption{Different divergence measures between $u$, the uniform distribution and $p$, a probability distribution over a Bernoulli random variable $X$. 
  Note that the \textbf{confidence penalty} is equivalent to $\KL(p \mid\mid u) = \Jzero$ and \textbf{label smoothing} is equivalent to $\KL(u \mid\mid p) =\Jone$ (see \cref{sec:family}). We include entropy $\ent(p)$ and $\mathrm{Eu}(u\mid \mid p) = \Jalphaf(u \mid\mid p)$ for $\alpha = 0.5$ and $G(p) = ||p||_2^2$. }
    \label{fig:div}
\end{figure}

We note $\Jalpha$ is equivalent to quadruple the Jensen--Shannon (JS) divergence and asymptotically approaches the Kullback--Leibler (KL) divergence for certain values of $\alpha$. Specifically, we have: 
\begin{align}
    \lim_{\alpha \to 0} \Jalpha (q \mid\mid \ptheta) &= \KL (\ptheta \mid\mid q) \\
    \lim_{\alpha \to 1} \Jalpha (q \mid\mid \ptheta) &= \KL (q \mid\mid \ptheta) \\
    \Jhalf (q \mid\mid \ptheta) &= 4\cdot \JS (q \mid\mid \ptheta)
\end{align}
We prove these relationships in \cref{app:to_kl} and \cref{app:to_JS}.
For ease, we define $\Jone \defeq \lim_{\alpha \to 1} \Jalpha$ and $\Jzero \defeq \lim_{\alpha \to 0} \Jalpha$.
We note the following two equivalences for these special cases.
\begin{prop}
$\nabtheta \Jone(u \mid\mid \ptheta) = \nabtheta \ent(q ,\ptheta)$. In words, the gradient of the loss with GER as $\alpha\!\rightarrow\!1$ is equivalent to the gradient of the loss augmented with \textbf{label smoothing}.
\end{prop}

\begin{prop}
$\nabtheta  \Jzero(u \mid\mid \ptheta) = \nabtheta \ent(\ptheta)$.  In words, the gradient of the loss with GER as $\alpha \rightarrow 0$ is equivalent to the gradient of the loss augmented with the \textbf{confidence penalty}. 
\end{prop}
\noindent See \cref{app:LS} and \cref{app:CER} for proofs.

\subsection{Formal Properties of $\Jalpha$}

When fitting a model $\ptheta$, we generally optimize the inclusive $\KL$, i.e. $\KL(\ptilde\mid\mid \ptheta)$, so that, among other reasons, $\ptheta$ has support everywhere that $\ptilde$ has support. 
However, it is unclear what relationships we want to encourage between the model $\ptheta$ and the uniform distribution $u$ during regularization as complete support of $u$ implies no word can ever have non-zero probability. 

Here we explore formal properties of $\Jalpha$ as a regularizer to gain insight into how, as a function of $\alpha$, these regularizers affect the learned distribution.

\paragraph{Magnitude.}
\cref{fig:div} shows the different divergence measures between $u$ and $\ptheta$. We see that $\Jone = \KL(u\mid\mid\ptheta)$ \textbf{(label smoothing)} is much larger than $\Jzero = \KL(\ptheta\mid\mid u)$ \textbf{(confidence penalty)} at values of $\ptheta$ farther from $u$. This indicates that $\Jone$ would be a stronger regularizer than $\mathrm{J}_{<1}$, i.e. penalize values of $\ptheta$ far from $u$ more heavily, given the same strength coefficient $\beta$. Note that it is not always the case that $\mathrm{J}_{<1} (u \mid\mid p) \leq \Jone(u \mid\mid p)$ for fixed $p$. We can, however, bound $\Jalpha$ from above and below by other quantities.

\begin{figure}[t!]
  
    \includegraphics[width=\textwidth]{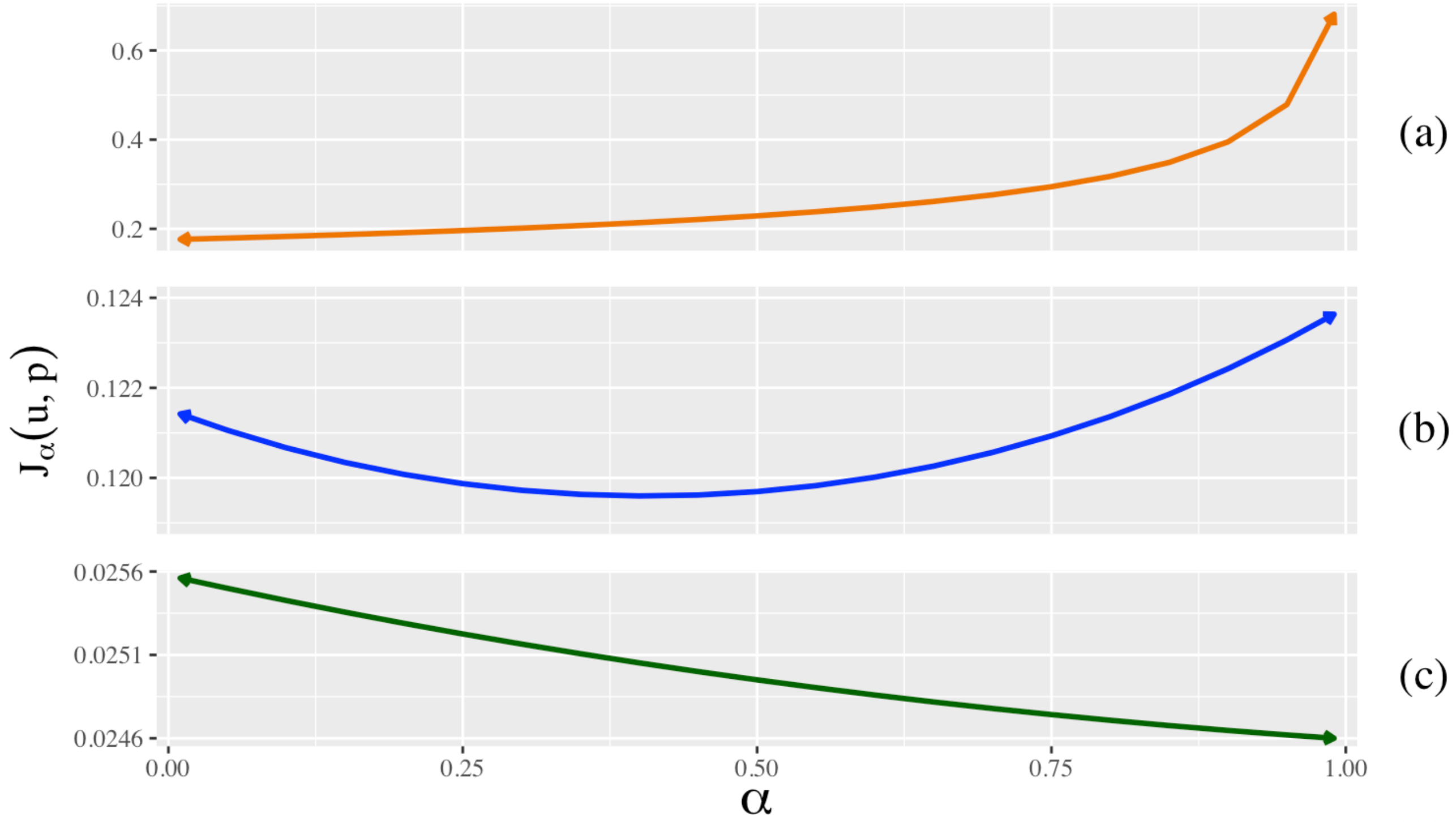}%
    
  \caption{$\Jalpha(u\mid\mid p)$ as a function of $\alpha$ for $u$, the uniform distribution, and $p$, a probability distribution over a 3-way categorical random variable, where for (a) $p = (0.0001,0.49995,0.49995)$ (b)  $p = (0.15,0.15,0.7)$ and (c)  $p = (0.25,0.25,0.5)$. There is no standard trend for $\Jalpha$ as purely a function of $\alpha \in (0,1)$.}
    \label{fig:alpha_lines}
\end{figure}

\begin{prop}\label{prop:not_monotonic}
The divergence $\Jalpha(u \mid\mid p)$ is \emph{not} a monotonic function of $\alpha$ for all distributions $p$. 
\end{prop}
\noindent A proof by counter example is shown in \cref{fig:alpha_lines}. 

\begin{prop}\label{prop:upper_bound}
For fixed $p$, $\Jalpha$ has bounds: \newline $0 \leq \Jalpha(u \mid\mid p) \leq \KL(u\mid\mid p) + \KL(p \mid\mid u)$.
\end{prop}
\noindent See \cref{app:bounds} for a proof.

\begin{table*}[!ht]
\ra{1.2}
  \centering
  \setlength\tabcolsep{4.3pt} 
  \footnotesize
  \adjustbox{width=\textwidth}{
  \begin{tabular}{ @{}llllllllllllll@{} }
   \toprule
      & \multicolumn{4}{c}{\bf WMT'14 De-En} & \multicolumn{4}{c}{\bf IWSLT'14 De-En} & \multicolumn{4}{c}{\bf MTTT Fr-En} \\
     & \textbf{$\alpha$} & \textbf{$\beta$} & \textbf{$\hat\ent$} & \bleu & \textbf{$\alpha$} & \textbf{$\beta$} & \textbf{$\hat\ent$} & \bleu & \textbf{$\alpha$} & \textbf{$\beta$} & \textbf{$\hat\ent$} & \bleu  \\
    \hline
    \it No Regularization         & -- & 0 & 0.11 & 31.1        & -- & 0 & 0.1 & 35.7                  & -- & 0 & 0.15 & 35.2   \\
    Label Smoothing $D_\Jone$ \tiny{($\gamma\!=\!0.1$)}     & 1 & 0.11& 0.23 & 31.3 \dd{0.2}      & 1 & 0.11 & 0.18 & 36.9 \dd{1.2}   & 1 & 0.11 & 0.18 &  36.5 \dd{0.8} \\
    \hdashline
    Label Smoothing $D_\Jone$       & 1 &0.35 & 0.38 &  31.7  \dd{0.6}    & 1& 0.50& 0.40 & 37.2 \dd{1.5}  & 1 &0.693 & 0.47 &  37.5 \dd{2.3}\\
    Confidence Penalty $D_\Jzero$   & 0 & 0.28 & 0.55 & 31.6 \dd{0.5}     & 0 & 0.76 & 0.81 & 37.5 \dd{1.8}   & 0 & 0.95 & 0.86 &  37.4  \dd{2.2} \\
    GER $D_\Jalpha$                 & 0.7 & 0.65 & 0.47 & 32.0 \dd{0.9}   & 0.5 & 1.00 & 0.56 & 37.5 \dd{1.8}   & 0.85 & 0.52 & 0.37 & 37.6  \dd{2.4} \\
    \bottomrule
  \end{tabular} }
  \caption{\bleu scores and normalized entropy $\hat\ent(\ptheta)$ on the test sets for WMT'14 De-En, WMT'14 De-En, and MTTT Fr-En. Results include baseline models with no (entropy) regularization and standard label smoothing with $\gamma\!=\!0.1$ (equivalent to $\beta \approx 0.11$). We report scores from the best model found (on validation set) for $D_\Jzero$, $D_\Jone$, and $D_\Jalpha$ over all $\alpha, \beta$ pairs. \bleu standard deviation across random seeds was typically $<0.1$ and always $<0.16$.\footnotemark \,Results for MTTT Ja-En and convolutional architectures can be found in \cref{app:res}. }
  \label{tab:mt_results}
\end{table*}

\paragraph{Sparsity.}\label{sec:sparsity}
Sparsity is generally a desirable trait in probabilistic models; specifically for structured prediction, it leads to improvements in performance and interpretability \cite{structured_sparsity, niculae2018}.
For example, \citet{sparsemax} showed the benefits of using sparsemax, which induces sparsity in an output distribution or attention layer, for natural language inference tasks. There are also intuitive reasons for allowing $\ptheta$ to be sparse. Part of modeling language generations tasks is learning when particular sequences \emph{cannot}, or at least should not, occur (e.g. are grammatically or syntactically incorrect). In these cases, a model should be able to assign 0 probability mass to that sequence.
However,  there is no sparse optimal solution $\ptheta$ when using label smoothing as the label smoothing loss function becomes divergent if $\ptheta$ does not assign probability mass $\forall y \in$ supp$(u)$.
\begin{prop}\label{prop:sparse}
$\Jalpha(u \mid\mid p)$ is finite for any $p \in \Omega$ and any $\alpha < 1$. As $\alpha \rightarrow 1$, $\Jalpha(u \mid\mid p)$ diverges iff $\exists y \in \mathrm{supp}(u)$ for which $p(y) = 0$.
\end{prop}
\noindent See \cref{app:sparsity} for a proof.

\footnotetext{We have $\alpha \approx 1$ as an exception; the standard deviation is slightly higher for larger values of $\beta$.}

\section{Experiments}\label{sec:exp}

We evaluate our family of entropy regularizers on two language generation tasks: machine translation and abstractive summarization. We then analyze trends in model performance as a function of $\alpha$ and model entropy\footnote{Model entropy is estimated as an average of the entropies of distributions at each time step during decoding, i.e. $\hat\ent(\ptheta) = D_H(\ptheta)$. Entropy is normalized by the maximum possible entropy for the given vocabulary size ($\log{|Y|}$) in all figures and tables to control for the fact that languages have vocabularies of different sizes.} and explore how this entropy affects other properties of language generation models. In the following experiments, each model is trained using \cref{eq:standard_reg} where $\reg(\vtheta) = D_\Jalpha(\ptilde \mid\mid \ptheta)$. We conduct searches over $\alpha$ and $\beta$ using Bayesian optimization \cite{Snoek:2012:PBO:2999325.2999464} to find the combination of regularizer $D_\Jalpha$ and strength coefficient $\beta$ that lead to the lowest loss on the development set for the respective task.\footnote{We only report results with generator function $G = -\ent$ as results using $G(z) = ||z||_2^2$ were consistently worse and often did not improve on the baseline; these results may be seen in \cref{app:res}.}  We additionally do a more fine-grained grid search over $\beta$ for $\Jzero$ (confidence penalty) and $\Jone$ (label smoothing) for completeness. All other model hyperparameters are held constant. We run experiments on multiple architectures and across several data sets to ensure trends are general.
\begin{figure*}
    \centering
    \includegraphics[width=1.0\textwidth]{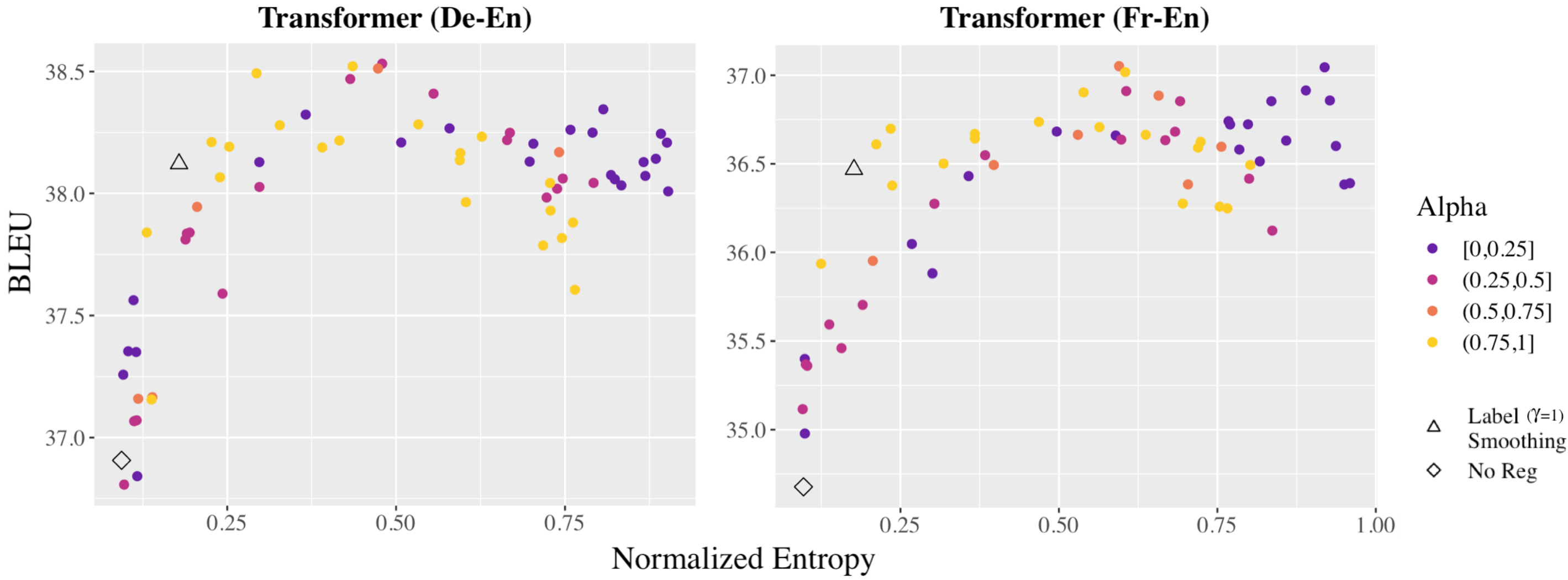}
    \caption{Model entropy $\hat\ent(\ptheta)$ vs. \bleu on IWSLT'14 German to English (De-En) and Multitarget TED Talks Task French to English (Fr-En) using a Transformer architecture; each point is a fully trained model, regularized with $D_\Jalpha$ for varying $\alpha$ and $\beta$. Label smoothing at standard $\gamma = 0.1$ and no (entropy) regularization are marked.}
    \label{fig:ent_v_bleu}
\end{figure*}  
\subsection{Neural Machine Translation}
We explore performance of the regularizer $D_\Jalpha$ on NMT systems using three language pairs and corpora of two different sizes on the following tasks: WMT'14 German-to-English (De-En) \cite{wmt14}, IWSLT'14 German-to-English (De-En) \cite{IWSLTbib}, and Multitarget TED Talks Task (MTTT) French-to-English (Fr–En) and Japanese-to-English (Ja-En) tasks \cite{dataset2}. 
For the larger WMT data set, we train fewer models using coarser-grained $\alpha$ and $\beta$ ranges. We perform experiments for both Transformers \cite{vaswani2017attention} and convolutional sequence-to-sequence models \cite{gehring2017convolutional}.\looseness=-1  

For reproducibility and comparability, we use the data pre-processing scripts provided by fairseq \cite{ott2019fairseq} and follow recommended hyperparameter settings from previous work \cite{vaswani2017attention, gehring2017convolutional} for baseline models. We use SacreBLEU \cite{sacrebleu} to calculate \bleu scores \cite{Papineni:2002:BMA:1073083.1073135}. Specific data pre-processing steps and model hyperparameter details are provided in \cref{app:data}. Decoding is performed with length-normalized beam search with a beam size of 5 unless otherwise stated. Early stopping was used during training; model parameters were taken from the checkpoint with the best validation set \bleu.

\begin{table}
\ra{1.2}
  \centering
  \footnotesize
  \begin{tabular}{ @{}lllll@{} }
   \toprule
   & $\alpha$ & $\beta$ & $\hat\ent(\ptheta)$ & \textsc{Rouge-L} \\
   \hline
   \it No Regularization & -- & -- & 0.08 & 40.5 \\
    Confidence Penalty $D_\Jzero$ & 0 & 0.15 & 0.19 & 40.9 \dd{0.4}\\
    Label Smoothing $D_\Jone$ & 1 & 0.1 & 0.2 & 40.9 \dd{0.4}\\
   GER $D_\Jalpha$ & 0.5 & 0.35 & 0.19 & 40.8 \dd{0.3} \\

     \bottomrule
  \end{tabular}
  \caption{\textsc{Rouge-L} on test set for CNN/DailyMail abstractive summarization task. Note that we replicate their reported result (achieved with label smoothing).}
  \label{tab:sum_results}
\end{table}

Results of our experiments are shown in \cref{tab:mt_results} and \cref{fig:ent_v_bleu}. We see the same relation between model entropy and \bleu with both Transformer and convolutional architectures and between different language pairs. We show results for the Transformer architectures inline as they are the current standard for many NLP tasks; results for convolutional architectures are in \cref{app:res}. 
Our results show better performance is achieved with values of $\alpha$ and $\beta$ other than those that correspond to label smoothing with $\gamma=0.1$, which is the commonly used value for the strength coefficient \cite{vaswani2017attention, edunov-etal-2018-understanding}. Moreover, the relationship between model entropy and evaluation performance is strong, following the same trend for all values of $\alpha$, which suggests tuning a model for a specific entropy rather than $\alpha, \beta$ may be a better method in practice. We discuss trends in \cref{sec:exp_alpha}.

\subsection{Abstractive Summarization}
We fine-tune BART \cite{lewis2019bart} on the CNN/DailyMail abstractive summarization task \cite{hermann2015teaching} with regularizer $D_\Jalpha$. Data pre-processing and other hyperparameter settings follow \newcite{lewis2019bart}. 
Results in \cref{tab:sum_results} show that optimal values of \textsc{Rouge-L}  \cite{lin-2004-rouge}, the evaluation metric, can be achieved by regularizing with $D_\Jalpha$ for different values of $\alpha$.
Notably, the entropy is virtually the same for the models that achieve top performance, demonstrating the closer relationship of performance with model entropy than with $\alpha$, discussed further in \cref{sec:exp_alpha}.

\subsection{Significance of $\alpha$ and Model Entropy}
\label{sec:exp_alpha}

We look at the strength of the relationship between the evaluation metrics and both $\alpha$ and the model's entropy. 
\cref{fig:ent_v_bleu} shows a quadratic relationship between model entropy and \bleu. On the other hand, the relationship between $\alpha$ (coloring of points) and \bleu is not an obvious one; the best performing models are regularized with various values of $\alpha$.

\begin{figure}
    \begin{minipage}[b]{\textwidth}
    \includegraphics[width=\textwidth]{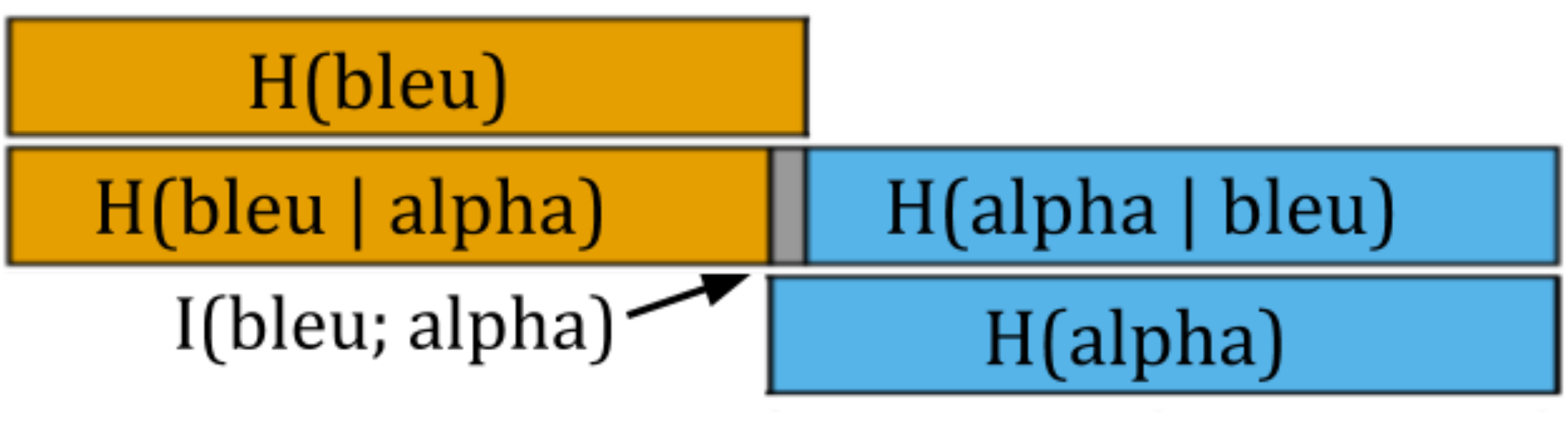}
   
  \end{minipage}
  \qquad
  \begin{minipage}[b]{\textwidth}
    \includegraphics[width=\textwidth]{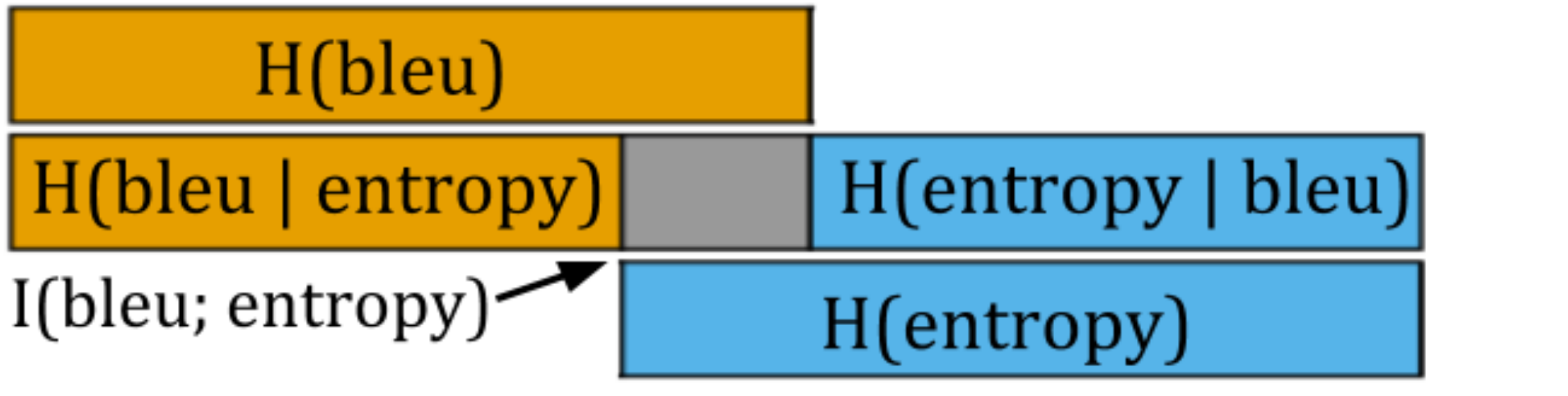}
    
  \end{minipage}
  \caption{Entropy $\ent(\cdot)$, Conditional Entropy $\ent(\cdot \mid \cdot)$ and Mutual Information $\mathrm{I}(\cdot ; \cdot)$ for \bleu with alpha ($\alpha$) and model entropy, respectively. Model entropy explains a greater portion of variability in \bleu than $\alpha$ does. Non-parametric estimates are used for all values \cite{nonpar_ent}. Data from IWSLT'14 De-En Transformer models. }
    \label{fig:mi}
\end{figure}

As correlation only tells us about linear relationships, we report mutual information to measure the strength of the relationship between $\alpha$, model entropy, and \bleu. Mutual information shows the proportion of entropy of a variable that is ``explained'' by another and is often used as a generalized correlation measure i.e. for nonlinear relationships \cite{correlation_measures}. We see in \cref{fig:mi} that model entropy has a much stronger relationship with \bleu than $\alpha$. Indeed, the normalized mutual information (NMI) between $\alpha$ and \bleu is $\approx 0.05$ compared to $\approx 0.25$ between model entropy and \bleu---implying that any flavor of entropy regularization can lead to similar performance.

While the relationship between $\alpha$ and \bleu is weak, it is still statistically significant. Some evidence for this exists in \cref{fig:ent_v_bleu} where a closer examination reveals that each level of $\alpha$ has a similar quadratic trend, albeit with a different offset. Specifically, the performance of models trained with $D_\Jalpha$ for $\alpha \in [0.75, 1]$ (which includes label smoothing) starts to degrade at lower levels of entropy than models trained with $D_\Jalpha$ for $\alpha \in [0, 0.25]$ (confidence penalty). As quantitative validation of this observation, we (i) run a conditional independence test to see whether \bleu and $\alpha$ are conditionally independent given model entropy and (ii) look at the range of $\beta$ for which $D_\Jalpha$ leads to good performance for different $\alpha$.

\paragraph{Conditional Independence.} If $\alpha$ and \bleu are conditionally independent it implies that the value of $\alpha$ does not supply any additional information about the value \bleu given model entropy, i.e. $\alpha$ does not matter when using the regularizer $D_\Jalpha$. We use a Monte Carlo permutation test where the null hypothesis is that no relationship between $\alpha$ and \bleu exists.\footnote{The underlying distributions of random variables are assumed to be Gaussian. See \citet{cond_test} for more details.} However, this test rejects the null hypothesis with $p$-value $< 0.05$, supporting the alternate hypothesis that $\alpha$ and \bleu are \emph{not} conditionally independent.

\begin{figure}
    \centering
    \includegraphics[width=\textwidth]{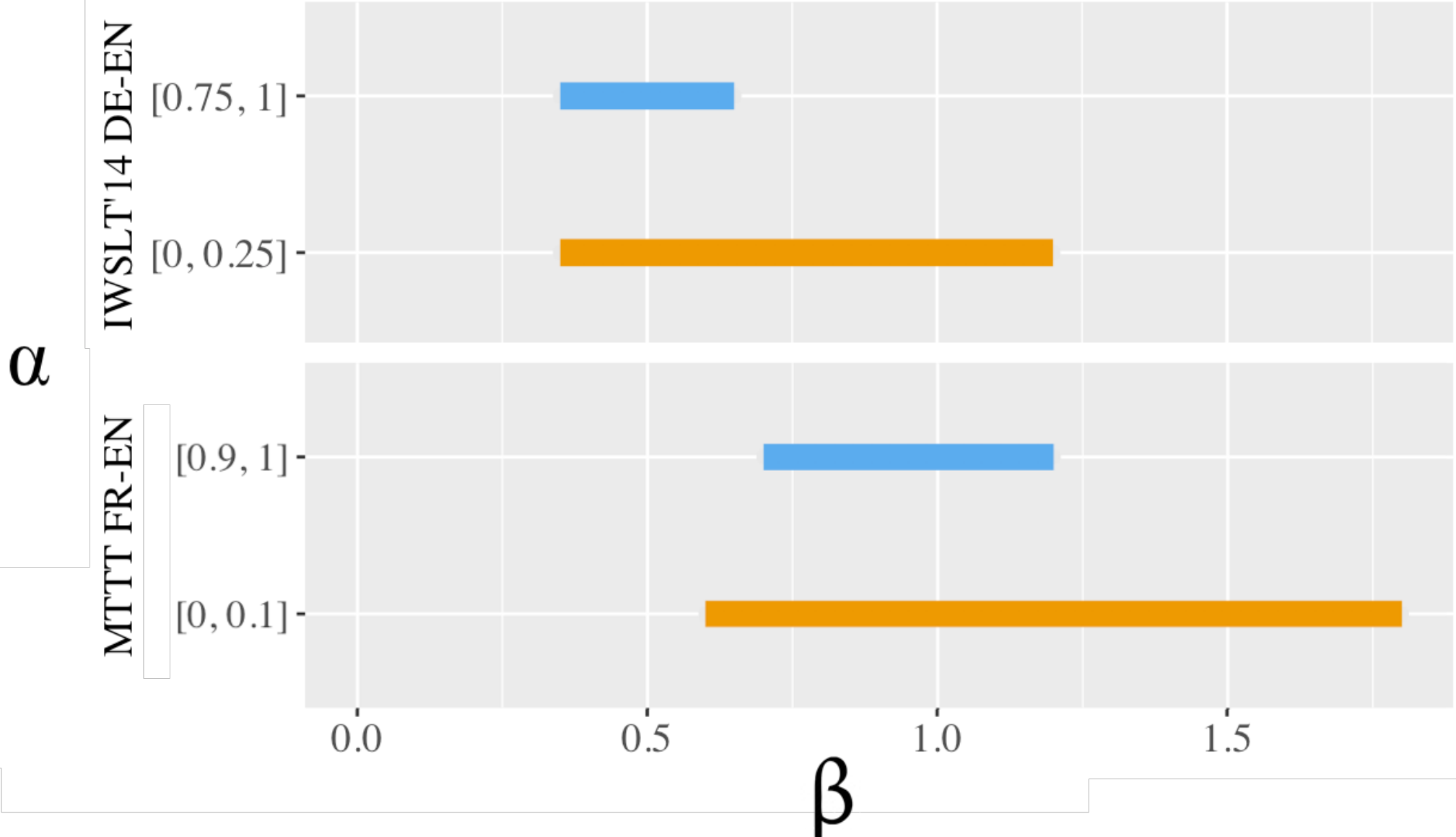}
    \caption{ Each line represents the range of $\beta$ for which $D_\Jalpha$ leads to performance within $\approx 1\%$ ($< 0.4$ \bleu) of the best overall model for the task. For $\alpha$ close to 1, (which includes label smoothing) $D_\Jalpha$ has a smaller optimal range, and so is harder to tune. }
    \label{fig:betas}
\end{figure}

\paragraph{Tuning $\beta$.} On the tasks for which we trained $> 60$ models, we take the subset of models for which performance is within $\approx 1\%$ ($< 0.4$ \bleu) of the best overall model. We then look at the range of $\beta$ used with the regularizer $D_\Jalpha$ for these models. The range of $\beta$ that meets the above criterion is much larger for $\alpha$ close to $0$ than for for $\alpha$ close to $1$ (see \cref{fig:betas}).
We contend this implies that $D_\Jalpha$ is easier to tune (i.e. it is more robust) for $\alpha \approx 0$ while for $\alpha \approx 1$, $D_\Jalpha$ is relatively sensitive to $\beta$.\looseness=-1

\begin{table}
\setlength\tabcolsep{5pt}
\begin{tabular}{lll}
\toprule
&  \multicolumn{2}{c}{\small Sparsity Threshold }\\
& \multicolumn{1}{c}{$e^{-10}$} & \multicolumn{1}{c}{$e^{-15}$} \\
\cmidrule(l){2-3}
\small

\footnotesize{Label Smoothing $D_\Jone$}   & \small$38\% \pm 0.01\%$  & \small$0.0\% \pm 5\text{e-}5\%$\\
\footnotesize{Confidence Penalty $D_\Jzero$} & \small$54\% \pm 5\text{e-}3\%$ & \small$0.7\% \pm 4\text{e-}4\%$ \\

\bottomrule
\end{tabular}
\caption{Percentage of words with $<\epsilon$ probability mass at different values of $\epsilon$ (below which we consider as functionally $0$) for models trained with $D_\Jone$ and $D_\Jzero$. To control for entropy, all models used in the calculation have entropy within the same $1\%$. }
\label{tab:sparsity}
\end{table}

\subsection{Sparsity}

We take a subset of models trained with regularizers $D_\Jzero$ and $D_\Jone$ and examine the sparsity of $\ptheta$. 
Results in \cref{tab:sparsity} support our formal analysis regarding the sparsity of $D_\Jzero$ and $D_\Jone$ in \cref{sec:sparsity}; 
$D_\Jone$ steeply penalizes sparsity while $D_\Jalpha$ for $\alpha < 1$ allows words to be assigned probability $\approx 0$.
\looseness=-1

\subsection{Sequence Likelihood}
\begin{figure}
    \centering
    \includegraphics[width=\textwidth]{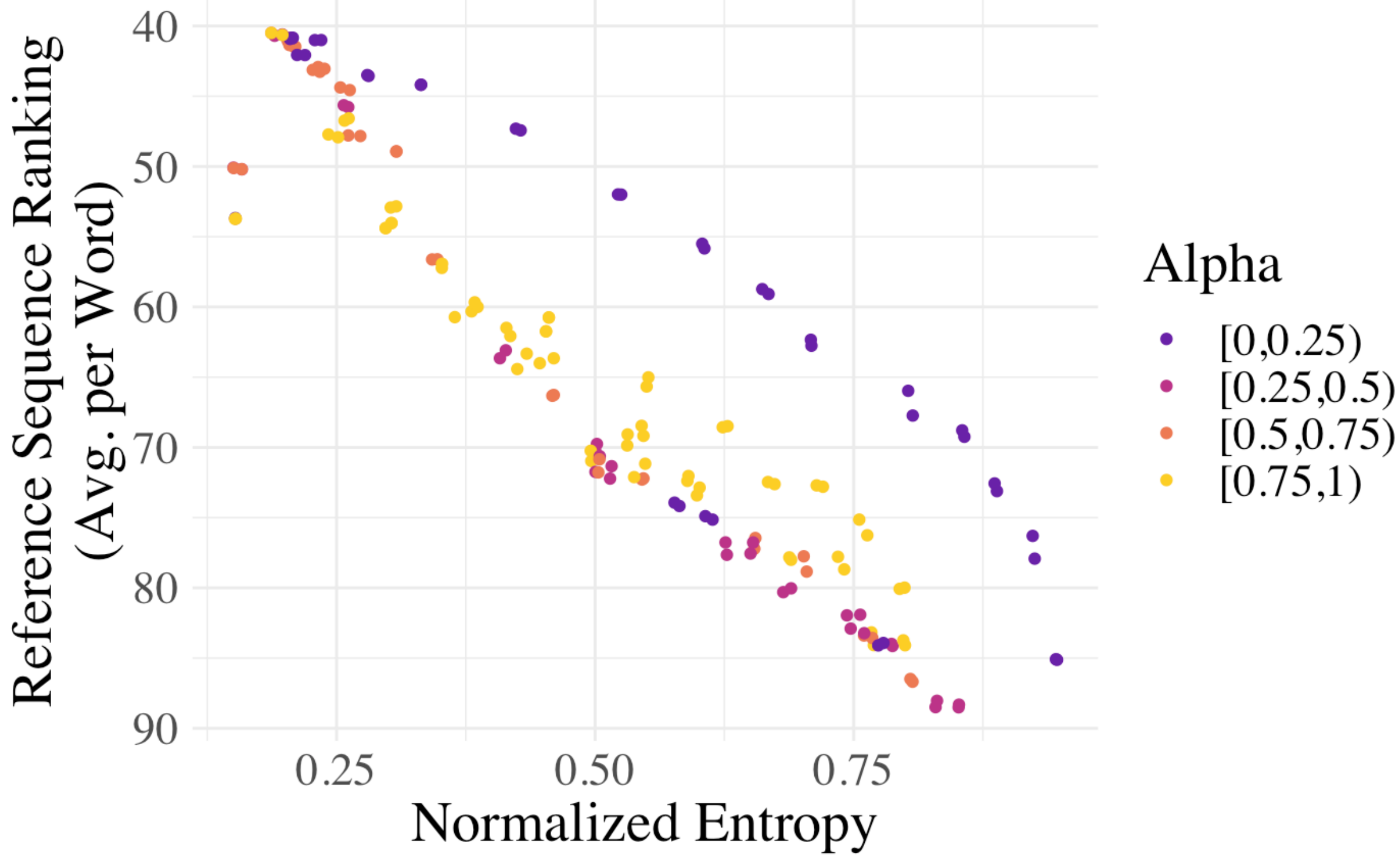}
    \caption{Average ranking in $\ptheta$ of words in the reference sequence  on the test set for IWSLT '14 (De-En) plotted against model entropy. Overall trends show a decrease in the ranking of the reference for models with more entropy regularization. Notably, the reference is generally ranked higher for models regularized with $D_\Jalpha$ for $\alpha \approx 0$ than for $\alpha \in [0.25,1)$.}
    \label{fig:seq_rank}
\end{figure}

We look at how the probability (under $\ptheta$) of the reference sequence on the test set changes with model entropy. While higher entropy in models trends positively with downstream evaluation metrics (\cref{fig:ent_v_bleu}), experiments show they often lead to lower log-likelihood of the reference sequence. Both of these observations have been made for models trained with label smoothing in previous works \cite{ott2018analyzing, when_does}. However, log-likelihood alone does not tell a complete story. During decoding, we search for the most probable sequence \emph{relative} to other candidate sequences. This implies that a more relevant calculation would be that of the overall ranking in $\calYs$ of the reference sequence or of the log-likelihood of the reference sequence relative to the most probable sequence. Since the former is typically impossible to calculate exactly due to the size of $\calYs$, we approximate it by looking at the average ranking in $Y$ of each word in the reference sequence.\looseness=-1

In \cref{fig:seq_rank}, we see that higher-entropy models generally rank the reference sequence lower than lower-entropy models; this result is surprising because higher-entropy models generally perform better on downstream evaluation metrics, e.g. \bleu. Notably, this decrease in ranking is less prominent for models regularized with $\alpha \approx 0$. In \cref{fig:seq_prob}, we see that while lower-entropy models place more probability mass on the reference sequence, the reference sequence is still far from probable compared to the decoded sequence. However, the ratio of log-likelihoods of the reference to the decoded sequence is larger for high-entropy models, which shows that, in this context, the reference sequence has higher \emph{relative} log-likelihood under higher-entropy models.\looseness=-1

\subsection{Decoding}

In language generation tasks, estimated distributions are fed to decoding algorithms to create sequence predictions. 
To fully understand how model entropy affects performance for these tasks, we must explore the potential interactions between model entropy and the decoding strategy. 

\begin{figure}[t!]
    \centering
    \includegraphics[width=7.5cm]{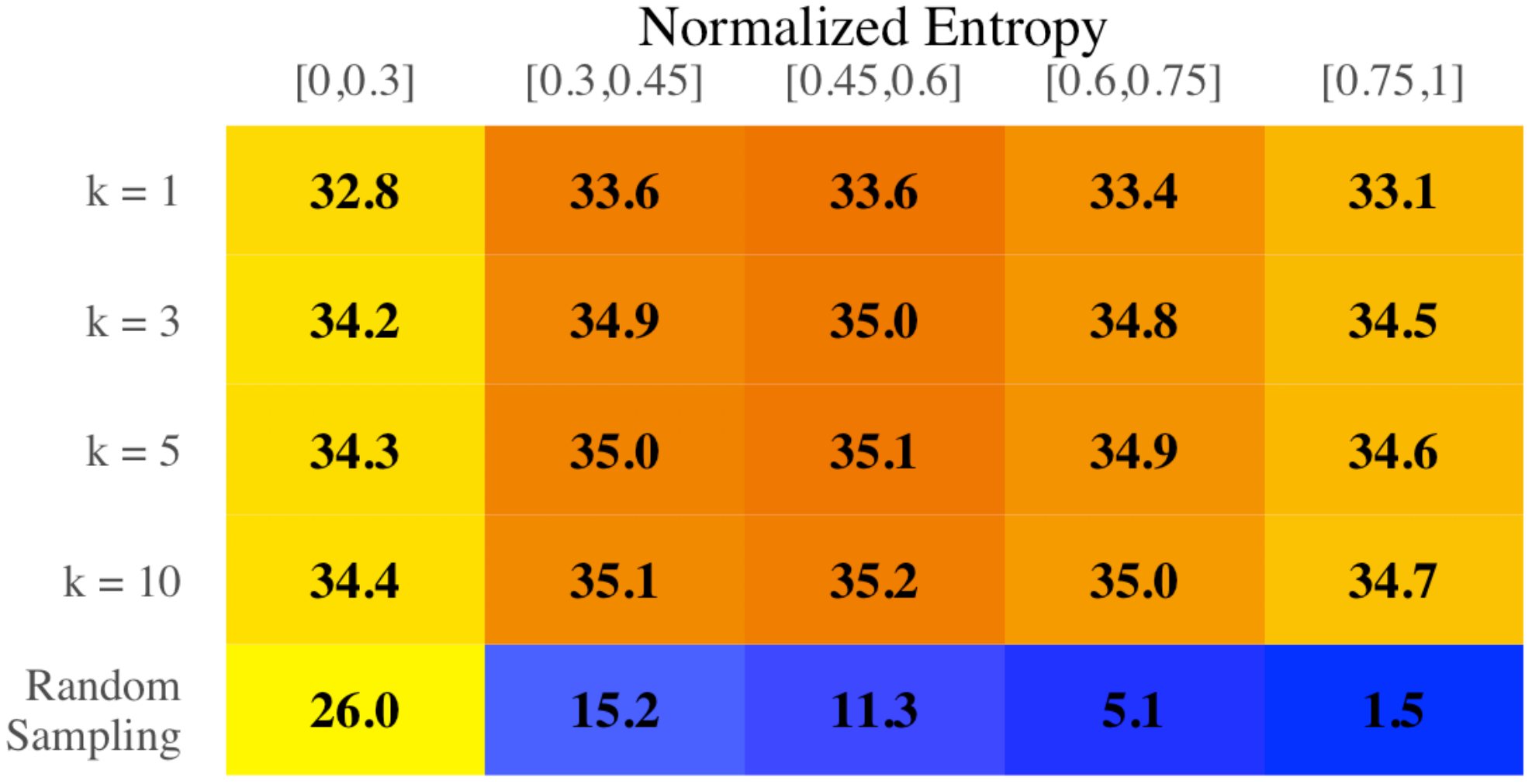}
    \caption{\bleu scores on IWSLT'14 De-En validation set with the convolutional architecture by decoding strategy and model entropy. The trend in \bleu stays remarkably constant for beam search as the beam width is varied. Performance declines drastically for higher entropy models when random sampling is used. Color reflects average distance from baseline model.}
    \label{fig:heatmap}
\end{figure}

\citet{decoding} saw that with label smoothing, prediction accuracy improved and so using a wider beam during beam search did not give further improvements; however, our results suggest otherwise. As shown in \cref{fig:heatmap}, the trend in \bleu vs. model entropy stays remarkably constant for beam search as the beam width is varied, including for greedy decoding (beam size of 1). Perhaps unsurprisingly though, higher entropy is detrimental to the performance of decoding with random sampling (with temperature $T=1$). However, this phenomenon could potentially be remedied by decreasing the temperature during decoding, a common practice for avoiding sampling from the tail of the distribution \cite{KirkGelaVecc83}.

\section{Discussion}

\begin{figure}
    \centering
    \includegraphics[width=0.83\textwidth]{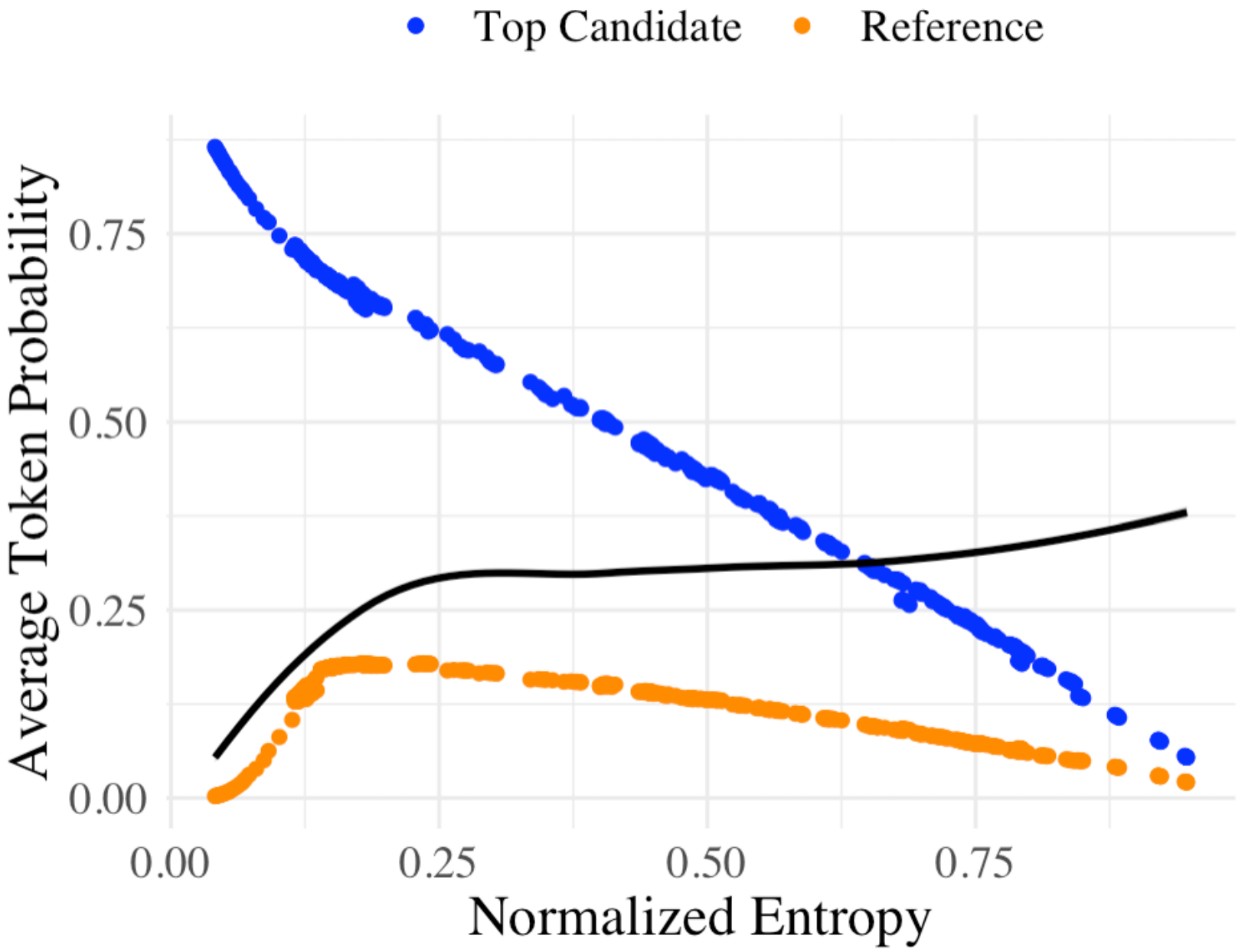}
    \caption{Average word probability of the reference and the most probable (for beam search with $k=5$) sequences plotted against model entropy on test set for IWSLT '14 (De-En). The black line is a smoothed estimate of their ratio.}
    \label{fig:seq_prob}
\end{figure}

Our experiments show entropy regularization has a number of beneficial  effects on natural language generation models. Clearly, low-entropy predictions, which are more aligned with the empirical distribution (\cref{fig:seq_prob}), are a sign of overfitting in a model since they lead to poor generalization abilities (\cref{fig:ent_v_bleu}). In other words, we observe that closely approximating the empirical distribution is at odds with a well calibrated model, i.e. a model $\ptheta(\yy \mid \xx)$ that matches the true, underlying probabilities $p(\yy \mid \xx)$.\footnote{This is different than the empirical distribution $\ptilde(\yy \mid \xx)$.} Entropy regularization appears to alleviate this problem; namely, for more regularized models, \cref{fig:ent_v_bleu} shows increased evaluation metric scores and \cref{fig:seq_prob} demonstrates an increase in the log-likelihood of the reference sequence relative to the highest probability sequence.\looseness=-1

\paragraph{Decoding.}

Overconfident predictions inhibit the ability to recover after a poor choice of words during decoding; \citet{decoding} suggest that higher-entropy models $\ptheta$, like the ones resulting from regularization with label smoothing, would alleviate this problem. 
Results throughout this paper support this hypothesis not just for label smoothing, but for the $D_\Jalpha$ family of entropy regularizers as well.

\paragraph{Choosing the baseline distribution.} \label{sec:dist}
Throughout this work, we use the uniform distribution $u$ as our \defn{baseline distribution} for the regularizer $D_\Jalpha$. However, one could also use some other distribution defined over the vocabulary such as the unigram \cite{decoding} or a function of word embedding distance with the target word \cite{kumar2018vmf, upgraded_LS}. Both have proven to be more effective than $u$ when used with label smoothing and the confidence penalty.
However, using distributions other than $u$ with $D_\Jalpha$ leads to indirect forms of entropy regularization. Specifically, the mathematical relationship to entropy regularization becomes more convoluted. Therefore, we leave the application of GER to other distributions as a topic for future work. 

\section{Related Work}\label{sec:rw}

Entropy regularization has a long history in reinforcement
learning \cite{Williams1991FunctionOU, pmlr-v48-mniha16, fox2015taming, SAC} where it has provided substantial improvements in exploration. Such methods have since been adapted for supervised learning where they have proven to be reliable forms of regularization for various probabilistic modeling tasks  \cite{NIPS2004_2740, smith-eisner-2007-bootstrapping}.

More recently, interpolating between exclusive and inclusive $\KL$ divergences has been explored in NMT by \citet{dual_skew}. However, this method was used for the objective function (i.e. between $\ptilde$ and $\ptheta$) and not as a regularization technique (i.e. between a baseline distribution $q$ and $\ptheta$). \citet{upgraded_LS} construct a baseline distribution $q$ as a function of word embedding distances to to use in place of the uniform distribution $u$ in the label smoothing equation. This work is complementary to ours, as $q$ can similarly be used in place of $u$ with GER. Finally, our work is closest to that of \citet{when_does}, which attempts to find the circumstances under which label smoothing has a positive effect on model performance. However, they do not explore entropy regularization on the whole nor do they attempt to provide an explanation for why label smoothing works. We attempt to answer the ``why'' question through a quantitative analysis of label smoothing and empirical exploration of the relationship between model entropy and performance.

\section{Conclusion}
We discuss the properties of generalized entropy regularization and provide empirical results on two language generation tasks. We find entropy regularization leads to improvements over baseline systems on evaluation metrics for all values of the parameter $\alpha$ with our regularizer $D_\Jalpha$. Theoretical and empirical evidence show label smoothing adds undesirable constraints to the model and is the hardest to tune of the regularizers tested. We therefore advocate
the use of alternate forms of entropy regularization for language generation tasks.

\bibliography{acl2019}
\bibliographystyle{acl_natbib}

\newpage
\clearpage 
\renewcommand{\thepage}{} 
\appendix

\onecolumn

\section{$\alpha$-Jensen to KL}\label{app:to_kl}

For reference, we repeat \cref{eq:def}, the definition of the skew Jensen divergence for some strictly convex function $G: \Omega \xrightarrow{} \mathbb{R}$ and probability distributions $p$, $q$:
\begin{align*}
    \Jalphaf (p \mid\mid q) \defeq \frac{1}{\alpha (1-\alpha) }\Big( (1-\alpha)G(p) + \alpha G(q) - \nabla G((1-\alpha)p + \alpha q) \Big)
\end{align*}

We can rewrite the $\alpha$-Jensen divergence with convex generator function $G$ in terms of the Bregman divergence
\begin{align*}
    \Jalphaf (p \mid\mid q) &= \frac{1}{\alpha (1-\alpha) }\Big( (1-\alpha)G(p) + \alpha G(q) - G(\spq) \Big) \\
    &= \frac{1}{\alpha (1-\alpha) }\Big( (1-\alpha)G(p) + \alpha G(q) - G(\spq)  \\ & \qquad
    \underbrace{- \alpha (1-\alpha) \langle p - q, \nabla G(\spq) \rangle -  \alpha (1-\alpha) \langle q - p, \nabla G(\spq) \rangle}_{\text{= 0, note $p-q$ in first inner product and $q-p$ in second}}   \Big)\\
    &= \frac{1}{\alpha (1-\alpha) }\Big( (1-\alpha)G(p) + \alpha G(q) - G(\spq) \\ &\qquad 
    - \underbrace{(1-\alpha) \langle \alpha(p-q), \nabla G(\spq) \rangle }_{\text{bring $\alpha$ inside the inner product since $b\langle v, w \rangle = \langle b \cdot v, w \rangle$}}  \\ & \qquad - \underbrace{\alpha  \langle(1- \alpha)(q-p), \nabla G(\spq) \rangle}_{\text{likewise, bring $(1 -\alpha)$ inside the inner product}}  \Big) \\
    &= \frac{1}{\alpha (1-\alpha) }\Big( (1-\alpha)G(p) + \alpha G(q) - G(\spq) \\ &\qquad 
    -(1-\alpha)\underbrace{ \langle p- (\spq), \nabla G(\spq) \rangle }_{\text{distribute $\alpha$ and rewrite}}  \\ & \qquad - \alpha  \underbrace{\langle q - (\spq), \nabla G(\spq) \rangle}_{\text{distribute $(1-\alpha)$ and rewrite}}  \Big) \\
    &= \frac{1}{\alpha (1-\alpha) }\Big( (1-\alpha)[G(p) - G(\spq) -
    \langle p- (\spq), \nabla G(\spq) \rangle] \\& \qquad \underbrace{  +  \alpha [G(q) -  G(\spq) - \langle q - (\spq), \nabla G(\spq) \rangle]}_{ \text{regroup terms based on multiplier (either $\alpha$ or $1 - \alpha$) so we can rewrite equation as two Bregman divergences}}  \Big)  \\
    &= \frac{1}{\alpha (1-\alpha) }\Big( (1-\alpha) D_G(p, \spq) +  \alpha D_G(q, \spq)  \Big) 
\end{align*}

We look at the behavior of $D_\Jalphaf (p \mid\mid q)$ as $\alpha \xrightarrow{} \{0, 1\}$  

\begin{align*}
    & \lim_{\alpha \xrightarrow{} 0} \frac{1}{\alpha (1-\alpha) }\Big( (1-\alpha) D_G(p, \spq) +  \alpha D_G(q, \spq)  \Big) \\
    &=  \lim_{\alpha \xrightarrow{} 0}\frac{1}{\alpha (1-\alpha) }\Big( (1-\alpha) \underbrace{D_G(p, p)}_{\text{ = 0}} +  \alpha D_G(q, p)  \Big) \\
    &= \lim_{\alpha \xrightarrow{} 0} \frac{1}{(1-\alpha) }D_G(q, p) \\
    &= D_G(q,p) 
\end{align*}

If we expand $D_G(q,p)$ using our generator function $G(p) = \sum_i p(i)\log p(i)$, we get
\begin{align*}
    &D_G(q,p) \\
    &= \sum_i q(i)\log q(i) - \sum_i p(i)\log p(i) - \langle q - p, log (p) - 1\rangle \\
    &= \sum_i q(i)\log q(i) - \sum_i p(i)\log p(i)  + \sum_i p(i)\log p(i)  - \sum_i q(i)\log p(i)\\ & \qquad   \underbrace{-\sum_i p(i) +  \sum_i q(i)}_{\text{ =0 since $q, p$ are both probability distributions summing to 1}} \\
    &= \sum_i q(i)\log q(i)  - \sum_i q(i)\log p(i) \\
    &= \KL (q\mid \mid p)
\end{align*}

Similarly, we can show $\lim_{\alpha \to 1} \Jalpha = \KL (p\mid \mid q)$

\section{$\alpha$-Jensen to Jensen--Shannon}\label{app:to_JS}

The proof that the $\alpha$-Jensen divergence is proportional to the Jensen--Shannon divergence is quite straightforward. If we evaluate $\Jalpha (p \mid\mid q)$ at $G = x\log x$ and $\alpha = \frac{1}{2}$

\begin{align*}
    \Jalpha (p \mid\mid q) &= \frac{1}{\alpha (1-\alpha) }\Big( (1-\alpha)G(p) + \alpha G(q) - G((1-\alpha)p + \alpha q) \Big) \\
    &= 4 \cdot \Big( \frac{1}{2} G(p) + \frac{1}{2} G(q) - G(\frac{1}{2}p + \frac{1}{2} q) \Big) \\
    &= 4 \cdot \Big( \frac{1}{2} p \log (p) + \frac{1}{2} q\log (q) - \frac{p+q}{2}\log (\frac{p+q}{2}) \Big)\\
    &= 4 \cdot \Big( \frac{1}{2} (p \log (p) -  p\log (\frac{p+q}{2})) + \frac{1}{2} (q\log (q) - q\log (\frac{p+q}{2})) \Big) \\
    &= 4 \cdot \Big( \frac{1}{2} \KL (p \mid\mid \frac{p+q}{2}) + \frac{1}{2} \KL (p \mid\mid \frac{p+q}{2}) \Big) \\
    &= 4 \cdot \JS(p \mid\mid q)
\end{align*}

\section{Label Smoothing}\label{app:LS}
For the case that $\alpha \rightarrow 1$, $p = u$, and $q = \ptheta$, we have
\begin{align*}
    \lim_{\alpha \to 1} \Jalpha(u \mid\mid \ptheta(\cdot \mid \xx)) &= \KL(\uu \mid\mid \ptheta) \\
    &= \sum_{y \in Y} \uu(y) \log \frac{\uu(y)}{\ptheta(y \mid \xx)} \\
    &= \sum_{y \in Y} \uu(y) \log \uu(y) - \sum_{y \in Y} \uu(y)\log\ptheta(y \mid \xx) \\
     &= \log |Y| - \sum_{y \in Y} \uu(y)\log\ptheta(y \mid \xx) \\
     &=  - \sum_{y \in Y} \uu(y)\log\ptheta(y \mid \xx) + N \\
\end{align*} 
\newpage

When $\Jone(u \mid\mid \ptheta(\cdot \mid \xx))$ is used as a regularizer for maximum likelihood training, we get the loss function 
\begin{align*}
    & \mathcal L(\theta) \\
    &=\KL\left(\ptilde(\cdot \mid \xx) \mid\mid \ptheta(\cdot \mid
    \xx)\right) + \beta \cdot \KL\left(\uu(\cdot) \mid\mid \ptheta(\cdot \mid
    \xx)\right) \\
    &=\underbrace{-\sum_{y \in Y} \Big( \ptilde(y \mid \xx) + \beta\cdot \uu(y) \Big) \log \ptheta(y \mid \xx)}_{\textit{unnormalized label-smoothed cross-entropy loss}} + N
\end{align*}

where $N$ is constant with respect to $\vtheta$.

\section{Classical Entropy Regularization} \label{app:CER}
For the case that $\alpha \rightarrow 0$, $p = u$, and $q = \ptheta$, we have

\begin{align*}
    \lim_{\alpha \to 0} \Jalpha (q \mid\mid \ptheta(\cdot \mid \xx)) &= \KL(\ptheta \mid\mid \uu) \\
    &= \sum_{y \in Y} \ptheta(y \mid \xx) \log \frac{\ptheta(y \mid \xx)}{\uu(y)} \\
    &= \sum_{y \in Y}\ptheta(y \mid \xx) \log \ptheta(y \mid \xx)  - \sum_{y \in Y} \ptheta(y \mid \xx) \log \uu(y) \\
    &= - \ent(\ptheta(y \mid \xx)) - \log \frac{1}{|Y|} \sum_{y \in Y} \ptheta(y \mid \xx)  \\
    &= - \ent(\ptheta(y \mid \xx)) - \log \frac{1}{|Y|} \\
    &= - \ent(\ptheta(y \mid \xx)) - N \\
\end{align*} 

When $\Jzero(u \mid\mid \ptheta(\cdot \mid \xx))$ is used as a regularizer for maximum likelihood training, we get the loss function 
\begin{align*}
    & \mathcal L(\theta) \\
    &=\KL\left(\ptilde(\cdot \mid \xx) \mid\mid \ptheta(\cdot \mid
    \xx)\right) + \beta \cdot \KL\left(\ptheta(\cdot \mid
    \xx \mid\mid \uu(\cdot))\right) \\
    &=\underbrace{\KL\left(\ptilde(\cdot \mid \xx) \mid\mid \ptheta(\cdot \mid \xx)\right) - \beta \cdot H(\ptheta(y \mid \xx)) }_{\textit{confidence penalty cross-entropy loss}} + \beta \cdot N
\end{align*}

\section{Bounds of $\Jalpha$}\label{app:bounds}

\textbf{Upper bound of $\Jalpha$:} \\

First note that $\KL(p \mid\mid q)$ is convex in $q$ when supp($p$) $\subseteq$ supp($q$), which must be true since $\q$ has support everywhere both $p$ and $u$ do for $\alpha \in (0,1)$. Therefore for $\alpha \in (0,1)$
\begin{align*}
    \KL(p \mid\mid \q) &\leq (1-\alpha)\KL(p\mid\mid u) + \alpha \KL(p\mid\mid p)\\
    &= (1-\alpha)\KL(p\mid\mid u)
\end{align*}
similarly,
\begin{align*}
    \KL(u \mid\mid \q) &\leq \alpha \KL(u\mid\mid p) 
\end{align*}

We then have:

\begin{align*}
    \Jalpha( u \mid \mid p) &= \frac{\alpha}{\alpha(1-\alpha)}\KL(p \mid\mid \q ) + \frac{1-\alpha}{\alpha(1-\alpha)}\KL(u \mid\mid \q) \\
    &\leq \frac{\alpha(1-\alpha)}{\alpha(1-\alpha)}\KL(p \mid\mid u ) + \frac{\alpha(1-\alpha)}{\alpha(1-\alpha)}\KL(u \mid\mid p) \\
    &= \KL(p \mid\mid u ) + \KL(u \mid\mid p)
\end{align*}

\textbf{Lower bound of $\Jalpha$:} \\

The bound from below is trivial given the definition of $\Jalpha$, however, it can more easily be seen by expressing $\Jalpha$ as the sum of $\KL$ divergences as above: 
\begin{equation*}
     \Jalpha( u \mid \mid p) = \frac{\alpha}{\alpha(1-\alpha)}\KL(p \mid\mid \q ) + \frac{1-\alpha}{\alpha(1-\alpha)}\KL(u \mid\mid \q)
\end{equation*}
\noindent Since $\alpha > 0$ and necessarily $\KL(\cdot \mid\mid \cdot) \geq 0$, the lower bound $0 \leq \Jalpha(u \mid\mid p)$ follows.

\section{No Sparse Solution for $\Jone$}\label{app:sparsity}
\begin{proof}
By definition, for any distribution $p$ over a vocabulary $Y$: 
\begin{equation}
   \Jone(u \mid\mid p) =  -\frac{1}{|Y|}\sum_{y \in Y} \log p(y) + \log |Y|
\end{equation}
\noindent Thus, if $\ptheta(y\mid \xx) \rightarrow 0$ for some $y \in Y$ and some $\xx \in \calX$, we have  $\Jone(u \mid\mid p) = \KL(u \mid\mid \ptheta) \rightarrow \infty$. This means that label smoothing enforces $\ptheta$ has support everywhere $u>0$, i.e. over all words $y \in Y$. For any $\alpha < 1$, $\Jalpha$ allows for sparse solutions since $\lim_{x \rightarrow 0} x\log x = 0$.
\end{proof}

\section{Data Pre-Processing and Hyperparameter Settings}\label{app:data}
For training with convolutional architectures we set hyperparameters, e.g. dropout, learning rate, etc., following \newcite{gehring2017convolutional}. On IWSLT'14 and MTTT tasks, we follow the recommended Transformer settings for IWSLT'14 in fairseq.\footnote{https://github.com/pytorch/fairseq/tree/master/examples/translation} 
Hyperparameters for models trained on the WMT task are set following version 3 of the Tensor2Tensor toolkit~\cite{tensor2tensor}. 
We use byte-pair encoding (BPE; \citealt{sennrich2016bpeacl}.) for all languages. Vocabulary sizes for WMT and IWSLT'14 are set from recommendations for the respective tasks in fairseq; for the MTTT tasks, vocabulary sizes are tuned on models with standard label smoothing regularization. 

Similarly, the CNN/DailyMail data set is pre-processed and uses BPE following the same steps as \cite{lewis2019bart}. Hyperparameters are the same as for their model fine-tuned on CNN/DailyMail. Details are available on the fairseq website.\footnote{https://github.com/pytorch/fairseq/blob/master/examples/bart/README.cnn.md}

\section{Additional Results}\label{app:res}

\begin{figure*}[htbp]
\RawFloats
    \begin{minipage}{0.46\textwidth}
    \includegraphics[width=\textwidth]{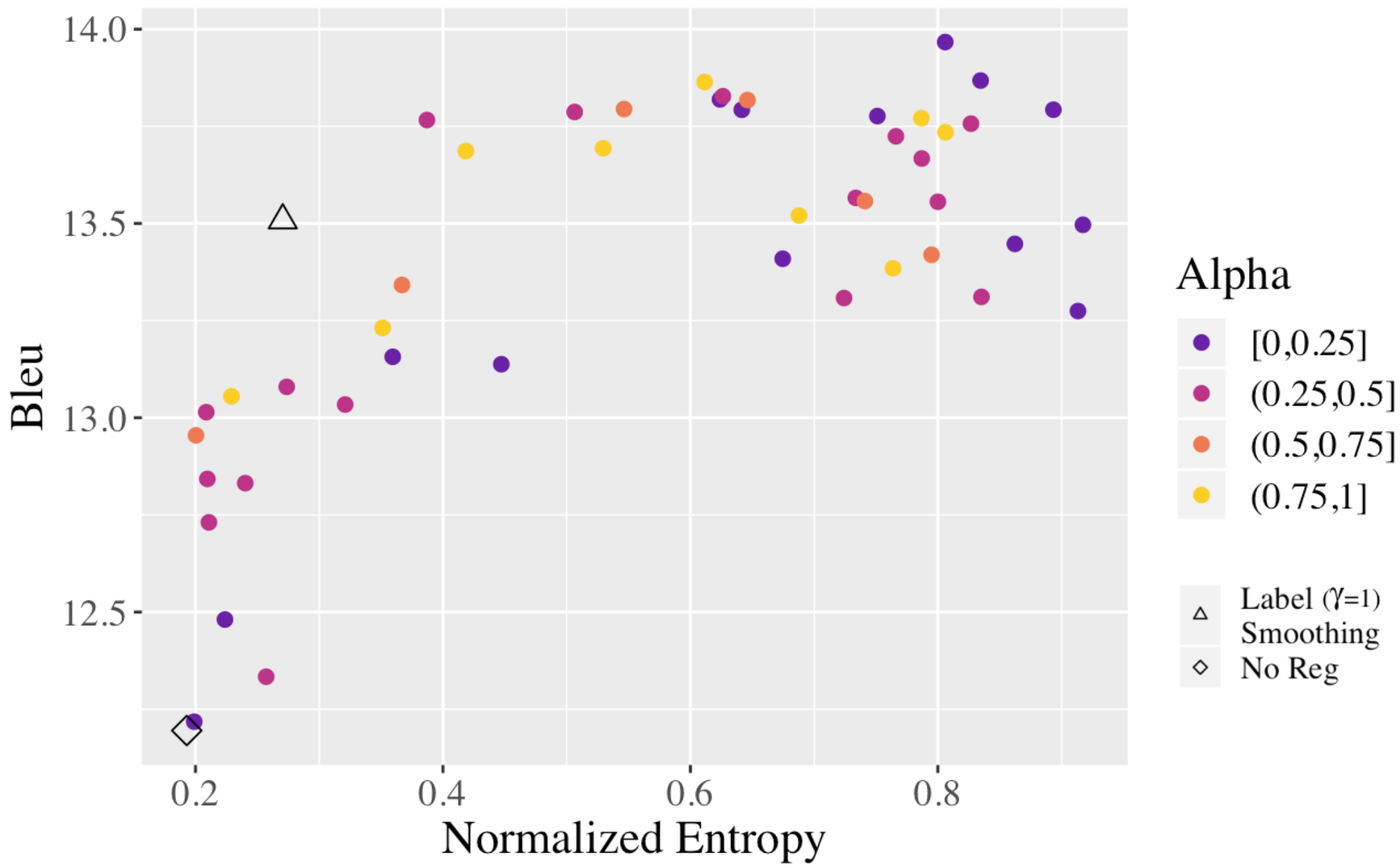}
   \caption{Model entropy vs. \bleu (validation set) on Multitarget Ted Talks Task Japanese to English (Ja-En) using a Transformer architecture; see Figure \ref{fig:ent_v_bleu} for additional information.}
  \end{minipage} \qquad
  \begin{minipage}{0.46\textwidth}
    \includegraphics[width=\textwidth]{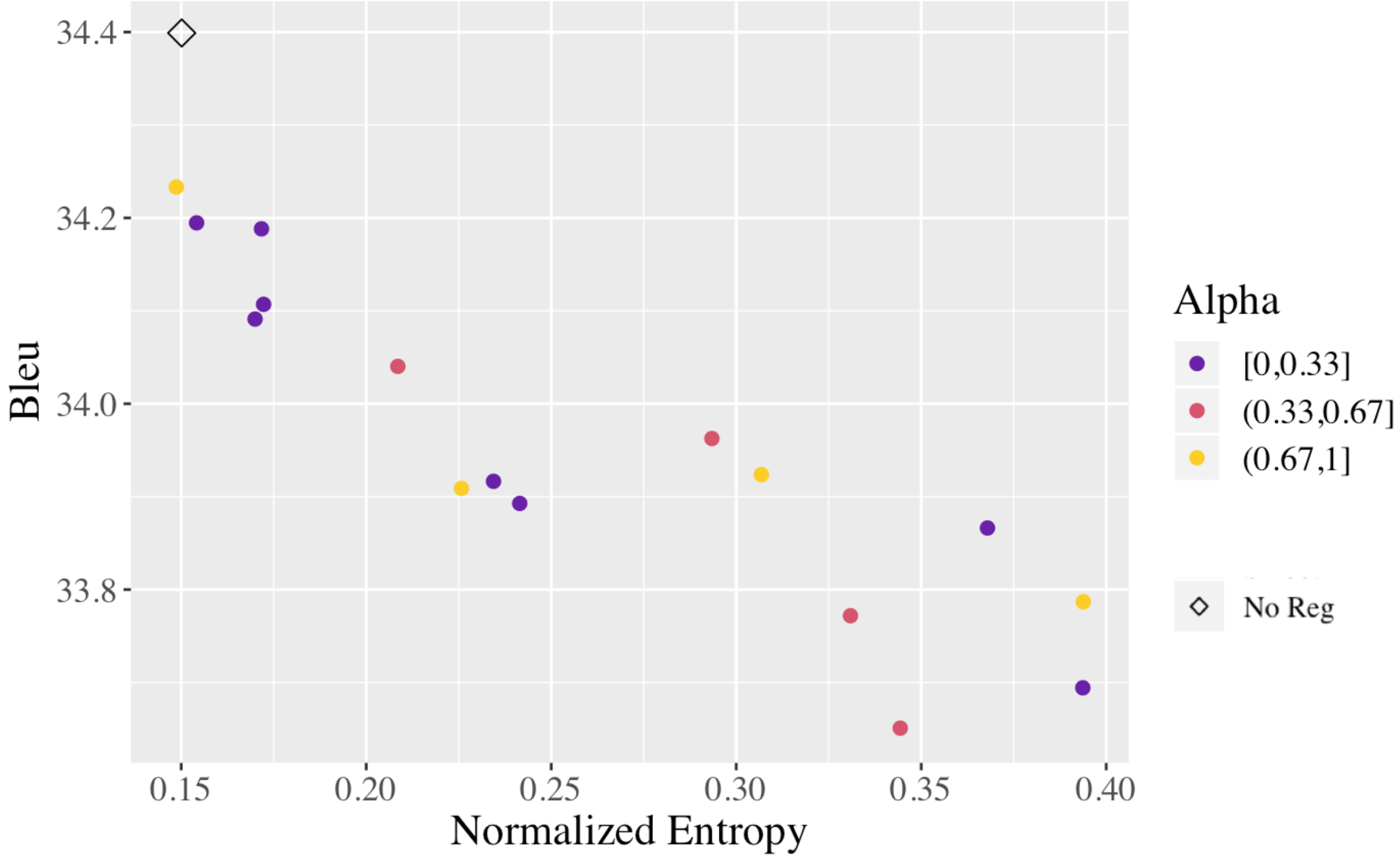}
    \caption{Model entropy vs. \bleu (validation set) on IWSLT'14 German to English (De-En) using a convolutional architecture and generator function $G(z) = ||z||_2^2$; see Figure \ref{fig:ent_v_bleu} for additional information.}
  \end{minipage}
 \end{figure*}

\begin{figure*}[t!]
    \centering
    \includegraphics[width=1.0\textwidth]{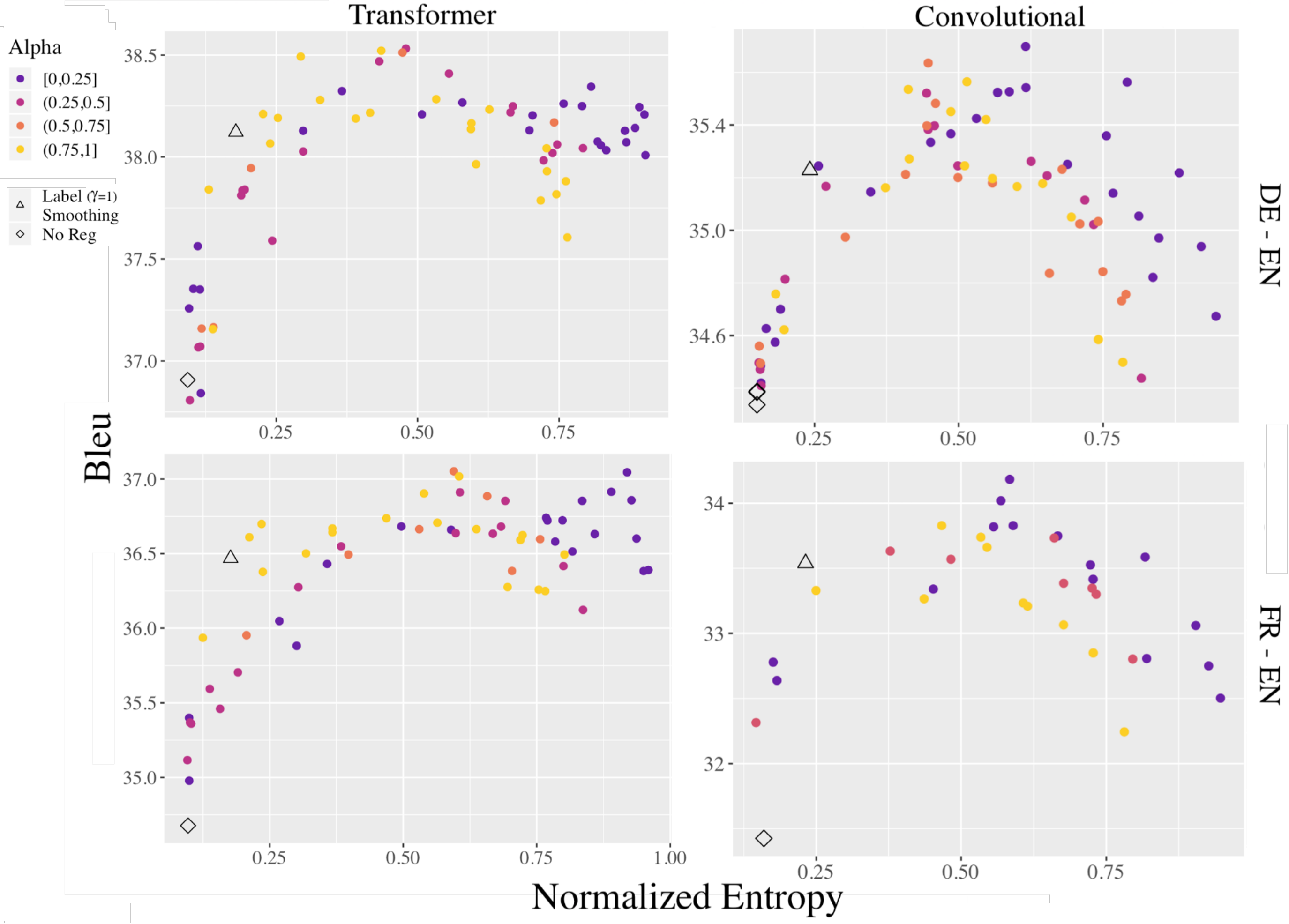}
    \caption{Model entropy vs. \bleu (validation set) on IWSLT'14 German to English (De-En) and Multitarget Ted Talks Task French to English (Fr-En) using Transformer and convolutional architectures; see Figure \ref{fig:ent_v_bleu} for additional information. }
    \label{fig:ent_v_bleu2}
\end{figure*}

\begin{table}
\ra{1.2}
  \centering
  \small
  \begin{tabular}{ @{}lllllllll@{} }
  \toprule
   & \multicolumn{4}{c}{\bf WMT'14 De-En (Convolutional)} & \multicolumn{4}{c}{\bf MTTT Ja-En (Transformer)} \\
  & \textbf{$\alpha$} & \textbf{$\beta$} & \textbf{$\hat\ent(\ptheta)$} & \bleu & \textbf{$\alpha$} & \textbf{$\beta$} & \textbf{$\hat\ent(\ptheta)$} & \bleu  \\
  \hline
  \it No Regularization &-	& 0 &0.15 & 33.2 &-	& 0 &0.19 & 13.8	\\
  Label Smoothing $D_\Jone$ \tiny{($\gamma=0.1$)} & 1	& 0.11 &0.25 & 34.1 \dd{0.9}& 1	& 0.11 &0.27 & 15.2 \dd{1.4}	\\
  \hdashline
  Label Smoothing $D_\Jone$ & 1	& 0.35 &0.42 & 34.6 \dd{1.4}& 1	& 0.96 &0.61 & 16.2 \dd{2.4}	\\
  Confidence Penalty $D_\Jzero$& 0	& 0.60& 0.79 & 34.7 \dd{1.5} &0	& 0.65 &0.80 & 15.9 \dd{2.1}	\\
 GER $D_\Jalpha$ & 0.75	& 0.60 & 0.45 & 34.8 \dd{1.6}& 0.42	& 1.7 & 0.76 & 15.9 \dd{2.1} \\
    
     \bottomrule
  \end{tabular}
  \caption{Test $\bleu$ for IWSLT'14 German-to-English using a convolutional architecture and for MTTT Japanese-to-English using a Transformer architecture; see \cref{tab:mt_results} for additional information.}
\end{table}\label{tab:de_results}
\end{document}